\documentclass[10pt,twocolumn,letterpaper]{article}
\usepackage{iccv}            
\usepackage[dvipsnames]{xcolor}
\definecolor{iccvblue}{rgb}{0.21,0.49,0.74}
\usepackage[pagebackref,breaklinks,colorlinks,allcolors=iccvblue]{hyperref}
\usepackage{adjustbox}
\usepackage{multirow}
\usepackage{subcaption}
\usepackage{graphicx}
\usepackage{makecell}
\usepackage{soul}
\usepackage{algorithm}
\usepackage{algpseudocode}
\usepackage[dvipsnames]{xcolor}
\usepackage{placeins}
\usepackage{listings}

\definecolor{my_green}{HTML}{008000}
\definecolor{my_purple}{HTML}{5c2e8a}
\definecolor{my_red}{HTML}{b10000}

\newcommand{\ours}{{\textnormal{\texttt{MonSTeR}}}\xspace}
\newcommand{\commas}{“}

\title{MonSTeR
: a Unified Model for Motion, Scene, Text Retrieval}

\author{%
  Luca Collorone\textsuperscript{1,*},
  Matteo Gioia\textsuperscript{1,*},
  Massimiliano Pappa\textsuperscript{1},\\
  Paolo Leoni\textsuperscript{1},
  Giovanni Ficarra\textsuperscript{1,3},
  Or Litany\textsuperscript{2},
  Indro Spinelli\textsuperscript{1},
  Fabio Galasso\textsuperscript{1}\\[0.6em]
  \textsuperscript{1}\,Sapienza University of Rome \qquad
  \textsuperscript{2}\,Technion, NVIDIA \qquad
  \textsuperscript{3}\,WSense \\
  {\tt\small \{name.surname\}@uniroma1.it} \quad
  {\tt\small or.litany@gmail.com}
}

\begin{document}
\maketitle
\begin{abstract}

Intention drives human movement in complex environments, but such movement can only happen if the surrounding context supports it. 
Despite the intuitive nature of this mechanism, existing research has not yet provided tools to evaluate the alignment between skeletal movement (\textbf{motion}), intention (\textbf{text}), and the surrounding context (\textbf{scene}).

In this work, we introduce \ours, the first MOtioN-Scene-TExt Retrieval model. Inspired by the modeling of higher-order relations, \ours constructs a unified latent space by leveraging unimodal and cross-modal representations.
This allows \ours to capture the intricate dependencies between modalities, enabling flexible but robust retrieval across various tasks.

Our results show that \ours outperforms trimodal models that rely solely on unimodal representations. Furthermore, we validate the alignment of our retrieval scores with human preferences through a dedicated user study. We demonstrate the versatility of \ours's latent space on zero-shot in-Scene Object Placement and Motion Captioning. Code and pre-trained models are available at \textcolor{Cerulean}{github.com/colloroneluca/MonSTeR}.
\end{abstract}    
\section{Introduction}
\label{sec:Intro}

\begin{quote}
    \textit{‘‘She asked me to stay and she told me to sit anywhere. But I looked around and I noticed there wasn't a chair."}
    \vspace{-1em}
    \begin{flushright}
        \textemdash \textit{ Norwegian Wood, The Beatles}
    \end{flushright}
\end{quote}

\begingroup
  \renewcommand\thefootnote{}
  \footnotetext{*\,Authors contributed equally.}
\endgroup

\begin{figure}[t] 
  \centering
  \includegraphics[width=0.5\textwidth]{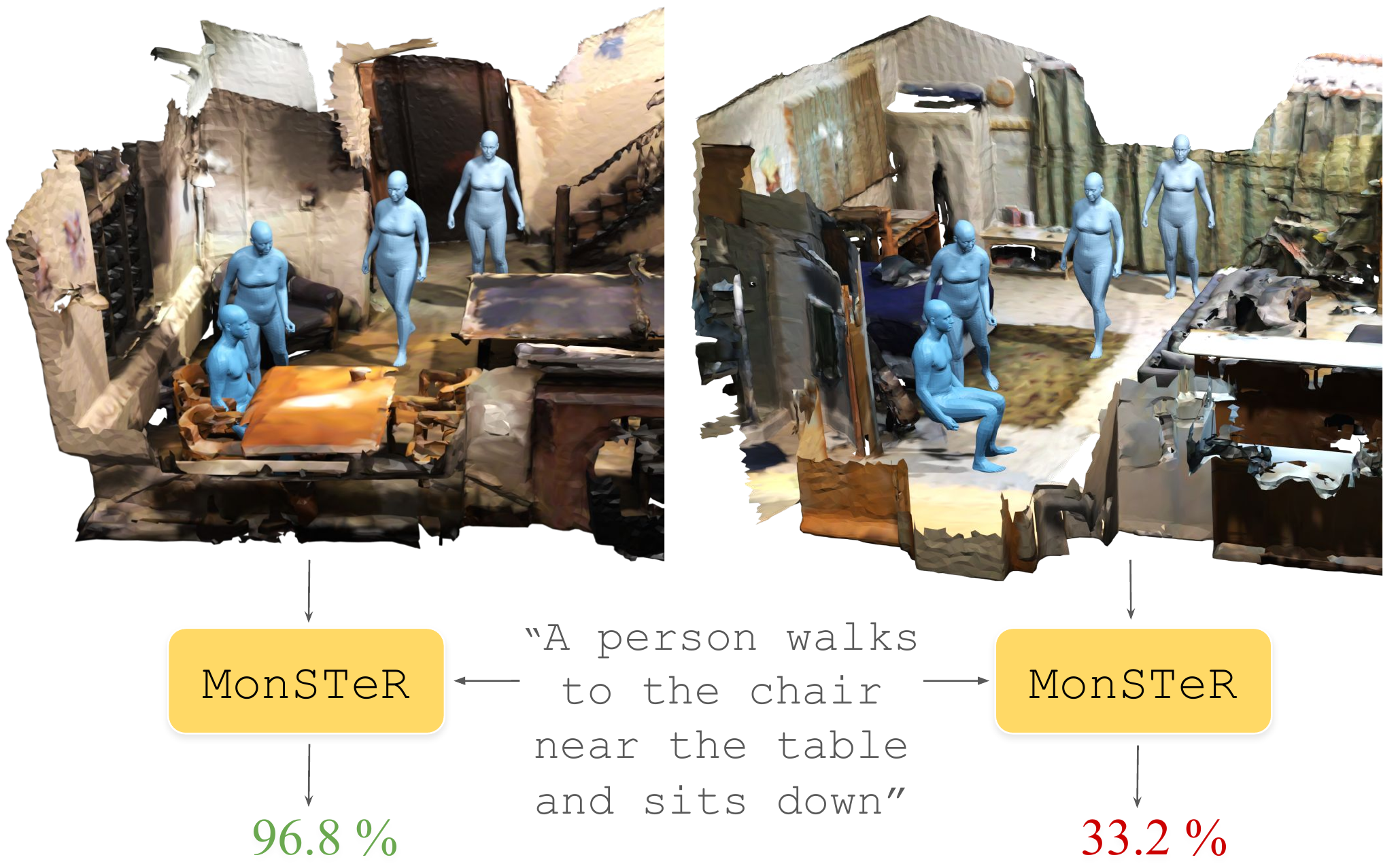}
  \caption{\ours can estimate coherence among text, motion, and scene by embedding them into a unified latent space. In the left image, all three modalities are coherent. However, in the right image, this coherence decreases, as there is no chair available in the scene.}
  \label{teaser}
\end{figure}

Humans can navigate and interact with environments by balancing their intentions with the opportunities presented by their surroundings. For instance, when the intended action is to \commas \textit{sit on a chair}" and the corresponding movement is performed, it would be perplexing if the environment lacked a chair (Fig~\ref{teaser}). This highlights the necessity for strong \textbf{coherence} and plausibility among intentions (text), movements (motion), and environments (scene) to create a scenario where human-scene interaction can naturally take place. 
Also, multiple captions could potentially describe the motion in Fig.~\ref{teaser} in isolation, but given the scene context, it is evident that the provided one is the most accurate. This confirms that exists a strong \textbf{interdependence} between these modalities. 

However, existing human motion generation and retrieval approaches are far from effectively leveraging this trimodal relation and fully capturing the interdependence between texts, motions and the environment in which they occur. In fact, recent retrieval models \cite{guo2022generating, petrovich23tmr, yu2024exploring, fujiwara2024chronologically} are unable to incorporate environmental context, causing motions to be depicted in an empty space and overlooking their intrinsic connection to the scene. 
Similarly, the methods for evaluating Human Scene Interaction models fail to provide a
global measure of coherence/realism \cite{wang2022humanise, wang2024move, cen2024generating}.
In fact, they try to obtain a score based on a number of different metrics, including person-scene interpenetration or distance to the target location. 
By contrast, we argue that these evaluations should take into account other critical aspects often neglected, such as the path taken, the plausibility of the motion, and the coherence among the three modalities.

Thus motivated, we introduce the first MOtioN-Scene-TExt Retrieval model, which we call \ours. This model is designed with a unified latent space that enables the assessment of coherence across three modalities: text $(t)$, motion $(m)$, and scene $(s)$.
Motivated by recent progress in topological deep learning \cite{papamarkou2024position}, we model inter-modal dependencies using higher-order interactions that go beyond pairwise relations. Unimodal terms, encoded by variational encoders, can be represented as nodes in a graph describing their interaction. In the same graph, we use cross-modal encoders to embed pairs of modalities, $ts$, $ms$, $mt$, representing its edges. To capture the higher-order interactions, we align unimodal and unimodal-bimodal representations: $(st,m)$ aligns the scene-text cross-modal representation with motion, $(mt,s)$ aligns the motion-text one with the scene, and $(ms,t)$ aligns the scene-motion one with text. The proposed modeling produces cohesive representations that deliver superior retrieval performance compared with approaches that align the three modalities using only unimodal terms. Furthermore, it provides the flexibility to perform various related tasks: retrieving a sample from one modality given another, a sample from one modality given a representation of two modalities, or vice versa. 

Additionally, we show that \ours can serve as an evaluator for text-conditioned Human Scene Interaction models. Specifically, we conduct (1) a dedicated user study confirming that its assessments align with human judgment, and (2) an evaluation revealing that \ours exhibits scene-motion grounding capabilities by assigning lower scores to motions that follow incorrect paths or interpenetrate the scene, confirming that our method can be used to assess path-scene plausibility.
Finally, we leverage the descriptiveness of the latent representations by performing Motion Captioning and zero-shot in-Scene Object Placement.

\noindent In summary, our contributions include:
\begin{itemize}
    \item A new retrieval model that explicitly models higher-order relations to unify text, motion, and scene, for the first time, within a unique latent space.
    \item A new method to ground the generations of text-conditioned Human Scene Interaction models within the scene and consequently evaluate their quality.
    \item Zero-shot evaluation of our model on the tasks of in-Scene Object Placement and evaluation on the downstream task of Motion Captioning.
\end{itemize}
\section{Related Works}
\label{sec:Related_Works}

This section introduces existing text-to-motion retrieval models, multimodal alignment, and aggregation techniques.

\subsection{Text-Motion Retrieval} 
Text-to-motion retrieval involves discriminating which motion sequence corresponds to a given text ($t2m$ task) and, conversely, identifying the correct text description for a query motion sequence ($m2t$ task). These tasks are often modeled using Contrastive Learning: a shared latent space is constructed between text and motion, where cosine similarity serves as a measure of coherence between the two modalities \cite{guo2022generating, tevet2022motionclip, petrovich23tmr, lbensabath2024, yu2024exploring, fujiwara2024chronologically}.
In particular, \cite{tevet2022motionclip} initiated this field by aligning texts, motions, and image renderings of the motions. Additionally, \cite{petrovich23tmr,lbensabath2024} proposed an effective and lightweight model to address this task and introduced several evaluation protocols for benchmarking. 
Existing approaches \cite{petrovich23tmr} represented motions through velocities and rotations in root space \cite{guo2022generating} or tackled this by encoding motions as image patches \cite{yu2024exploring} processed via a pre-trained Vision Transformer (ViT).
Unlike previous work, we offer a latent space that can estimate coherence not only between text and motion but also with the scene itself.

\subsection{Beyond Two-Modal Alignment} Recent works extending beyond text and motion have proposed various strategies for aligning multiple modalities within a shared latent space. Specifically, \cite{girdhar2023imagebind, xue2023ulip, xue2024ulip, lei2024vit, liu2023open} align all modalities to one modality (or a subset) chosen as a reference.
Alternatively, \cite{mai2022hybrid, Ruan_2024_WACV, yin2024tri}  align each modality with all the others, creating intricate interactions among modalities. Particularly relevant to our task, \cite{yin2024tri} presents a trimodal learning framework that incorporates human-centric videos, motion, and textual instructions.

In this work, we explore both approaches. Since \ours is designed to harness the interdependence between modalities, it is trained by aligning individual modality representations and paired modalities, achieving an all-to-all alignment. We validate our approach through an ablation study, aligning scene and motion solely with text, showing degraded performance compared to \ours.

\subsection{Modality Aggregation} 
Early approaches used separate encoders for each modality, combining them via latent averaging \cite{girdhar2023imagebind, Ruan_2024_WACV}, or adopted conditional generation—e.g., generating scenes from motion \cite{fouhey2012people, brooks2022hallucinatingposecompatiblescenes, nie2022pose2roomunderstanding3dscenes} or motion from scenes \cite{puttingpeople2023, lee2023locomotionactionmanipulationsynthesizinghumansceneinteractions}. However, these approaches do not explicitly model modality interdependence in the latent space, limiting the ability to learn joint representations. 
Other strategies, such as those in \cite{gong2022contrastive, chen2024scen}, process all modalities through a shared encoder to exploit cross-modal interactions, yet still produce separate embeddings per modality. This prevents the model from representing multiple modalities jointly or retrieving a single modality from a composite latent representation.
Additionally, \cite{poklukar2022geometric} aligns global (multi-modal) and per-modality encoders, but lacks the flexibility to encode partial modality combinations into a unified representation. Finally, \cite{delmas2024poseembroider} encodes pose, image, and instruction together into a unified embedding and can reconstruct the original modalities. In contrast, \ours promotes cross-modal learning by explicitly combining unimodal and pairwise cross-modal encoders, which provides richer representations that span all the available modalities.

\section{MonSTeR}

In this section, we present the design of \ours. We start with data representation, then detail the high-order decomposition that informed our latent space, and conclude with the chosen optimization objective.
\label{sec:Methodology}

\subsection{Data Representation} 
Following ~\cite{petrovich23tmr}, we use DistilBERT~\cite{sanh2020distilbertdistilledversionbert} to obtain an initial representation for text $t$, in the form of a 768-dimensional feature vector, i.e.\ $t\in\mathbb{R}^{768}$.
As in \cite{3dvista,jia2024sceneverse}, the scene $s$ is treated as a colored point cloud of $N$ points containing objects with semantic information, i.e.\ $s\in\mathbb{R}^{N\times6}$, where the six features are the concatenation of the $(x,y,z)$ coordinates and the RGB colors. We represent the 3D human motion $m$ as vectors $m\in\mathbb{R}^{T\times3\times22}$, where $T$ is the number of frames, 3 the $x,y,z$ coordinates of joints, and 22 the number of joints. 
\subsection{Designing MonSTeR's latent space}
\begin{figure}[t] 
    \centering
    \includegraphics[width=0.45\textwidth]{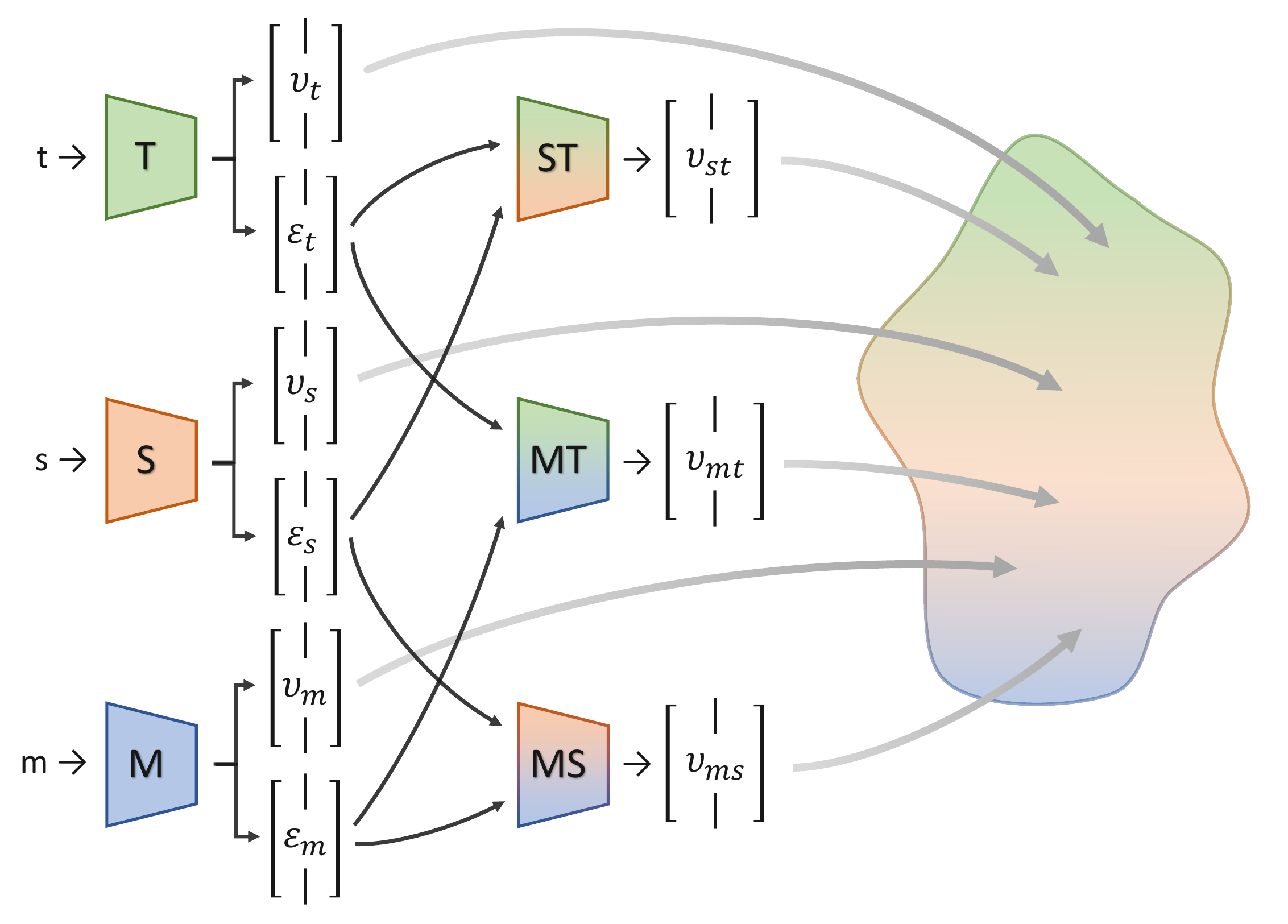} 
    \caption{\ours's Architecture Overview. Each input modality $t,s,m$ is processed by its single-modality encoder. From the first output tokens of $T,M,S$ we sample vectors $v_{t}$, $v_{m}$, and $v_{s}$. The remaining tokens of each encoder's output, namely $\varepsilon_t$, $\varepsilon_m$, $\varepsilon_s$, are pairwise concatenated and passed through cross-modal encoders to generate joint latent vectors ($v_{st}$, $v_{mt}$, $v_{ms}$).}
    \label{fig:architecture}
\end{figure}

Aligning scene, motion, and text into a unified multimodal latent manifold is a challenging goal.
The complexity arises from the need to capture the many-to-many relationship among these modalities: a single text may correspond to various motions, and the appropriate motion may differ depending on the scene context. The triplet (scene, motion, text) defines an event with minimal ambiguity\footnote{Some ambiguity still exists, as underlying intentions and personal preferences influence human movement}. Unlike pairwise alignments, this triplet captures higher-order interactions between the modalities, providing a richer and more coherent representation of the underlying event \cite{petrovich23tmr}.

\noindent\textbf{Higher-Order Relations in Topological Spaces.}
Topology provides a robust framework to analyze and extract insights from data \cite{bodnar2021wlt, papamarkou2024position}. Therefore, we set to exploit this theory to describe the process of embedding text, scene, and motion in a unified latent space via contrastive learning. This requires modeling \textit{part-whole} relations and it is therefore different from other contrastive objectives targeting pairwise relation between unimodal terms. Turning to topological geometry, this three-way relation represented by a polygonal face $\mathcal{P} =\{tsm\}$ can be decomposed into the composition of the three edges that are a \textit{part} of the triangle $\mathcal{E} = \{ts,sm,mt\}$. Similarly, each edge can be decomposed into the vertices that define it, resulting in a set of vertices $\mathcal{V} = \{t,s,m\}$. By aligning unimodal terms, we encode their pairwise relationship in the latent space. On the other hand, by aligning an edge with the opposite vertex, we encode their higher-order relation. All in all, topology \cite{papillon2024architecturestopologicaldeeplearning} demonstrates that representing the three-way relation of ($s,m,t$) requires representing the single modes $\{s,m,t\}$ and the pairwise cross-modal terms $\{sm,mt,ts\}$. We verify this empirically, cf.~Sec.~\ref{sec:ablations}. 

\noindent\textbf{Contrastive Learning to Learn High Order Relations.}
We align unimodal to unimodal and unimodal to cross-modal terms via contrastive learning. The goal is to embed this trimodal interaction in the latent space.
Following the schematics of Fig.~\ref{fig:architecture}, we encode the unimodal terms $\{t,m,s\}$ using transformer-based variational autoencoders. The output of each encoder is split into two: (1) the first two tokens are interpreted as the mean and log-variance of the latent distribution, and they are used to sample latent vectors $v_{t}$, $v_{m}$, and $v_{s}$. (2) The rest of these tokens, namely $\varepsilon_t$, $\varepsilon_m$, $\varepsilon_s$, are pairwise concatenated and processed by cross-modal encoders (e.g. $\varepsilon_s$ and $\varepsilon_t$ are concatenated to obtain the input of the $ST$ encoder). Each one of the three cross-modal encoders outputs mean and log-variance used to sample the cross-modal latent vectors $v_{st}$, $v_{mt}$, and $v_{ms}$. This process and the architecture of its encoders are  further detailed in the Supplementary Material.

\subsection{Training MonSTeR}

\label{sec:monster}
\begin{figure}[t] 
    \centering
    \includegraphics[width=0.43\textwidth]{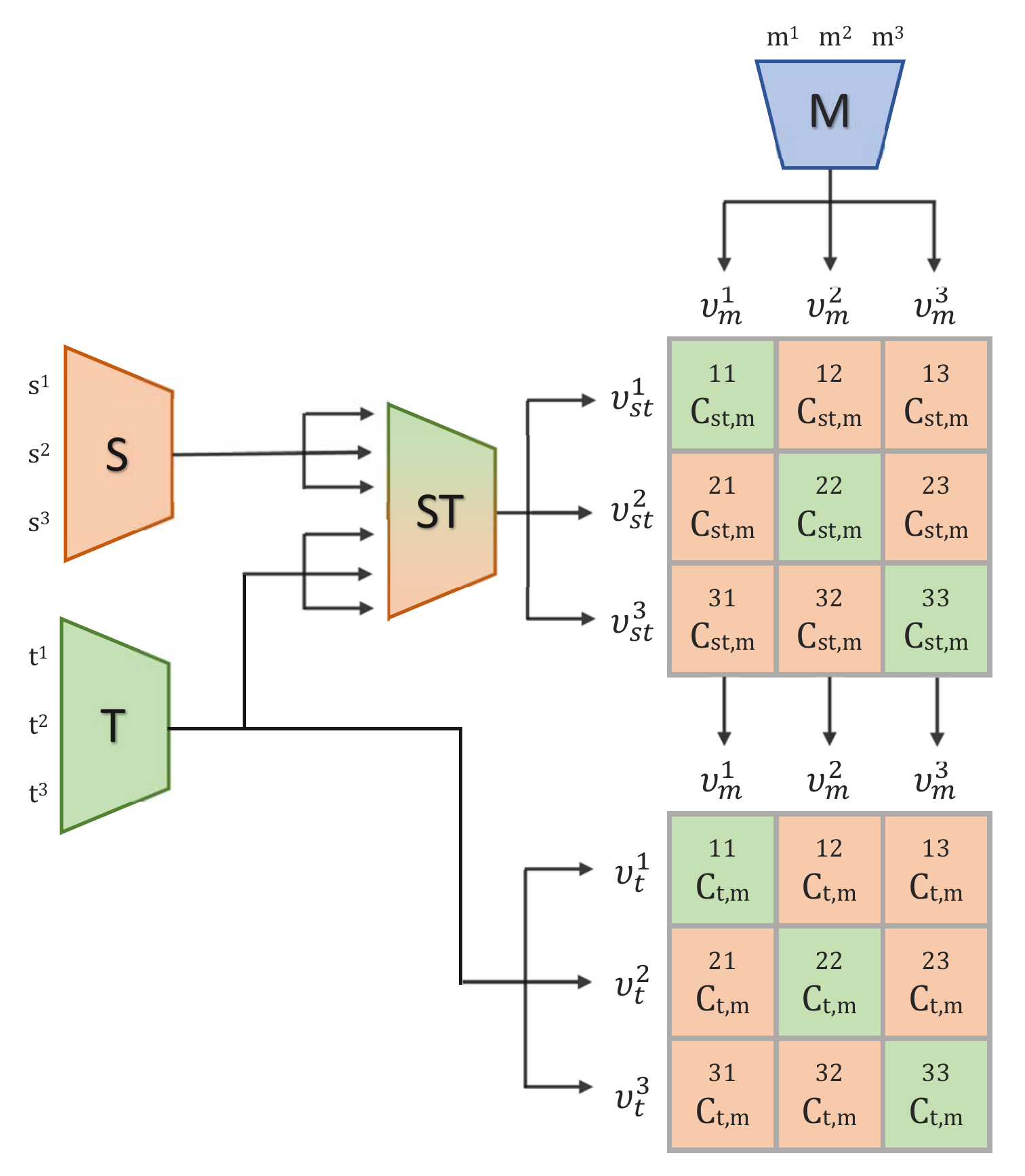} 
    \caption{Composition of $C_{st,m}$ and $C_{t,m}$ similarity matrices. For $C_{st,m}$, motion latents $v_{m}^{n}$ are compared with the cross-modal scene-text latents $v_{st}^{n}$, while for $C_{t,m}$, they are compared only with text latents $v_{t}^{n}$. Green cells are the locations of similarity scores between positive samples. Our optimization objective promotes assigning higher scores to these cells in all matrices $C_{i,j}$.}
    \label{fig:loss}
\end{figure}

Contrastive objectives are about maximizing the similarity between positive pairs, or different modalities representing the same object. In our objective, we include all the terms needed to recover unimodal and cross-modal relations of the three modalities. We exclude terms that could encourage the model to ignore one of the modalities during alignment, leading to degenerate solutions, e.g., $(st,t)$ or $(st,s)$. This is to prevent a collapse in the cross-modal encoders, that could simply learn an identity function. This leaves us with the resulting terms for the contrastive objective:

\begin{center}
    $K=\{(t,s), (m,t), (m,s), (st,m), (mt,s), (ms,t)\}$
\end{center}

Once we collect all the latent vectors for all the modalities in a batch, we proceed with the alignment procedure.  
Formally, for a batch of $N$ tuples of latent vectors 
\resizebox{\linewidth}{!}{$(v^1_t, v^1_m, v^1_s, v^1_{st}, v^1_{mt}, v^1_{ms}), \cdots, (v^N_t, v^N_m, v^N_s, v^N_{st}, v^N_{mt}, v^N_{ms})$}, we compute the $N \times N$ cosine similarity matrices $C_{i,j}$ for each pair of unimodal or cross-modal vectors $(i, j) \in K$. Positive pairs appear on the main diagonal, while the off-diagonal terms represent negative pairs.
For clarity, Fig.~\ref{fig:loss} illustrates the composition process of two example similarity matrices: $C_{st,m}$ which aligns unimodal to cross-modal embeddings and $C_{t,m}$ which relies solely on unimodal ones. 
Each similarity matrix is then used to compute the \mbox{InfoNCE} loss \cite{oord2018representation}, which we aggregate in the final loss:
\[
 \mathcal{L}_{\text{tot}} =  \frac{1}{|K|} \sum\limits_{\substack{(i,j) \in K}} \frac{\mathcal{L}_{\text{NCE}}(C_{i,j})}{N}
\]
This process pushes the model to bring the positive vectors (those belonging to the same tuple) closer while distancing them from other vectors in the batch.

\noindent\textbf{Cross-Modal Retrieval.} 
The unified latent space enables retrieval of scenes, motions, and texts, as well as their combinations, using either single modalities or by specifying two of them. For instance, to retrieve the correct motion given a scene and text (\textit{st2m} task), we compute the scene-text cross-modal embedding using the $ST$ encoder in Fig~\ref{fig:architecture}, and then calculate its cosine similarity with the outputs of the $M$ encoder.
\section{Experiments}
\label{sec:Experiments}

In this section, we describe the evaluation datasets, retrieval protocols, and \ours{} ablations, as well as its application to Human-Scene Interaction motion generation. We also perform In-Scene Object Placement and Motion Captioning tasks to demonstrate the effectiveness of the learned representations.

\subsection{Datasets}

To benchmark \ours, we use two recent datasets, \cite{wang2022humanise, jiang2024scaling}. We select these datasets as they are large-scale options that feature a substantial variety of motions across numerous scenes, all accompanied by textual annotations.

\noindent\textbf{HUMANISE+.} HUMANISE \cite{wang2022humanise} is a synthetic dataset containing 19.6K text-annotated human motion carefully aligned in 643 3D scenes.

The text descriptions in this dataset primarily reference an object of interaction, omitting details about its spatial location, attributes, and other objects in the scene. When multiple instances of the object exist, the descriptions fail to specify which one is relevant to the associated motions. To address this limitation, we use LLAMA3 \cite{dubey2024llama} to recaption the original texts to include references to nearby objects and express spatial context around the motion path. This approach substantially enhances the grounding of text descriptions within the scene. A detailed description of this process is provided in the Supplementary Material. We refer to this dataset as HUMANISE+. 

\noindent\textbf{TRUMANS+.} TRUMANS \cite{jiang2024scaling} comprises 15 hours of motion-captured data representing human interacting with 100 varied indoor scenes and 20 distinct object types. While TRUMANS texts' describe the motion accurately, their connection to the scene is not strong, primarily focusing on the object of interaction. Furthermore, the 3D scans of the environment are significantly larger than those in HUMANISE, often containing irrelevant elements such as rooms unrelated to the motion. Although a recaptioning process identical to that of HUMANISE+ was not feasible due to the lack of detailed scene segmentation labels, we applied a similar strategy to strengthen the alignment between text, motion, and scene content. We refer to this dataset as TRUMANS+.

\begin{table*}[t]
\centering
\renewcommand{\arraystretch}{1.0}
\small
\caption{Tasks and protocols' results on HUMANISE+ \cite{wang2022humanise} test set. Reported metrics are mRecall computed across ranks \{1,2,3,5,10\}. Greyed-out results are not directly comparable with those in the same column, as they do not leverage scene information.}
\begin{adjustbox}{max width=\textwidth}
\renewcommand{\arraystretch}{1.00}
\begin{tabular}{l|l|c|c|c|c|c|c||c|c|c|c|c|c||c}
\hline
\textbf{Protocol} & \textbf{Method}  & \textit{st2m}               & \textit{m2st}   & \textit{ms2t}       & \textit{t2sm}   & \textit{tm2s}      & \textit{s2mt}      & \textit{t2m}   & \textit{m2t}   & \textit{s2m}        & \textit{m2s}       & \textit{t2s}   & \textit{s2t} & avg.   \\
\hline
\multirow{5}{*}{All}
& TMR \cite{petrovich23tmr}    & \textcolor{gray}{2.73}             & –               & \textcolor{gray}{2.05} & –               & –                   & –                   & 2.73 & 2.05 & –                   & –                  & –               & –     & -          \\
& MoPa \cite{yu2024exploring}  & \textcolor{gray}{\underline{4.49}}  & –               & \textcolor{gray}{3.53} & –               & –                   & –                   & \textbf{4.49}    & \textbf{3.53} & –                   & –                  & –               & –    & -           \\
& TMR + S                & 4.10                                & \underline{3.30} & \underline{5.81}      & \underline{4.79} & 1.08                & 1.98                & 3.40             & \underline{3.22}            & \underline{1.41}    & 1.02               & \underline{1.24} & \underline{1.38} & \underline{2.72}\\
& MoPa + S               & 2.10  & 2.45  & 1.62  & 1.94  & \underline{3.28} & \underline{3.06} & 1.15 & 0.70 & \textbf{1.86} & \textbf{2.21} & 1.07 & 1.19 & 1.88 \\
& \ours                 & \textbf{13.91}                      & \textbf{13.14}  & \textbf{8.46}         & \textbf{10.39}  & \textbf{4.09}       & \textbf{4.45}       & \underline{3.62}             & 3.11            & 1.12                & \underline{1.16}   & \textbf{1.68}   & \textbf{2.51} & \textbf{4.80}  \\
\hline
\multirow{5}{*}{\makecell{Small \\ Batches}}
& TMR \cite{petrovich23tmr}    & \textcolor{gray}{56.0} & –               & \textcolor{gray}{56.01} & –               & –                   & –                   & 56.0 & 56.01 & –                   & –                  & –               & – & -              \\
& MoPa \cite{yu2024exploring}  & \underline{\textcolor{gray}{61.12}}            & –               & \textcolor{gray}{61.39} & –               & –                   & –                   & \underline{61.12}            & \underline{61.39}           & –                   & –    & -              & –               & –               \\
& TMR + S                & 57.08                               & \underline{55.49}           & \underline{70.77}      & \underline{67.05} & 33.86               & 37.26               & 59.42            & 58.17           & 27.78               & 25.73              & \underline{38.31} & \underline{38.45} & \underline{47.44} \\
& MoPa + S               & 48.83 & 40.73 & 40.46 & 39.23 & \underline{48.37} & \underline{40.73} & 28.47 & 27.00 & \textbf{36.80} & \textbf{36.66} & 35.17 & 37.45 & 38.32 \\
& \ours                 & \textbf{79.15}                      & \textbf{79.01}  & \textbf{76.30}        & \textbf{77.42}  & \textbf{57.55}      & \textbf{57.06}      & \textbf{63.20}   & \textbf{61.89} & \underline{34.11}   & \underline{34.47}  & \textbf{50.18}  & \textbf{49.76}  & \textbf{60.00}\\
\hline
\end{tabular}
\end{adjustbox}
\label{tab:retrieval_humanise}
\end{table*}
\begin{table*}[t]
\centering
\renewcommand{\arraystretch}{1.0} 
\small 
\caption{Tasks and protocols' results on TRUMANS+ \cite{jiang2024scaling} test set. Reported metrics are mRecall computed across ranks \{1, 2, 3, 5, 10\}. Greyed-out results are not directly comparable with those in the same column, as they do not leverage scene information.}
\begin{adjustbox}{max width=\textwidth}
\renewcommand{\arraystretch}{1.00}
\begin{tabular}{l|l|c|c|c|c|c|c||c|c|c|c|c|c||c}
\hline
\textbf{Protocol} & \textbf{Method} & \textit{st2m} & \textit{m2st} & \textit{ms2t} & \textit{t2sm} & \textit{tm2s} & \textit{s2mt} & \textit{t2m} & \textit{m2t} & \textit{s2m} & \textit{m2s} & \textit{t2s} & \textit{s2t} & avg. \\
\hline
\multirow{5}{*}{All}
& TMR \cite{petrovich23tmr}   
    & \textcolor{gray}{\textbf{17.55}} 
    & – 
    & \textcolor{gray}{\textbf{17.71}} 
    & – 
    & – 
    & – 
    & \textbf{17.55} 
    & \textbf{17.71} 
    & – 
    & – 
    & – 
    & – 
    & -
    \\
& MoPa \cite{yu2024exploring} 
    & \textcolor{gray}{4.22} 
    & – 
    & \textcolor{gray}{6.58} 
    & – 
    & – 
    & – 
    & 4.22 
    & 6.58 
    & – 
    & – 
    & – 
    & – 
    & -
    \\
& TMR + S              
    & 4.04 & 4.44 & \underline{9.01} & \underline{4.67} & 9.14 & 1.61 & 5.65 & \underline{12.03} & 1.11 & 5.89 & 6.74 & \underline{1.65} & \underline{5.49}
    \\
& MoPa + S             
    & 4.83  
    & \underline{5.20}  
    & 4.11  
    & 4.39  
    & \textbf{13.20}  
    & \underline{2.14}  
    & 3.52  
    & 3.85  
    & \textbf{2.06}  
    & \textbf{10.79}  
    & \underline{8.30}  
    & 1.04 & 5.28  \\
& \ours              
    & \underline{10.54} & \textbf{10.29} & 8.59 & \textbf{8.51} & \underline{12.71} & \textbf{3.94} & \underline{5.67} & 5.90 & \underline{1.94} & \underline{7.88} & \textbf{10.35} & \textbf{2.80} & \textbf{7.42} \\
\hline
\multirow{5}{*}{\makecell{Small \\ Batches}}
& TMR \cite{petrovich23tmr}   
    & \textbf{\textcolor{gray}{82.2}} 
    & – 
    & \textbf{\textcolor{gray}{82.13}} 
    & – 
    & – 
    & – 
    & \textbf{82.2} 
    & \textbf{82.13} 
    & – 
    & – 
    & – 
    & – 
    & -
    \\
& MoPa \cite{yu2024exploring} 
    & \textcolor{gray}{48.55} 
    & – 
    & \textcolor{gray}{56.44} 
    & – 
    & – 
    & – 
    & 48.55
    & \underline{56.44} 
    & – 
    & – 
    & – 
    & – 
    & -
    \\
& TMR + S &    
48.46 & \underline{47.63} & 54.05 & 51.85 & 30.38 & 26.7 & \underline{53.38} & 55.12 & 23.24 & 25.27 & 30.06 & 25.37 & 39.29
    \\
& MoPa + S              
    & 42.40 
    & 44.84 
    & 52.76
    & \underline{52.99} 
    & \underline{46.50} 
    & \textbf{43.66} 
    & 42.20 
    & 42.35 
    & \textbf{39.41} 
    & \textbf{43.62} 
    & \underline{31.59} 
    & \underline{29.12}
    & \underline{42.60}\\
& \ours                
& \underline{60.8} & \textbf{61.4} & \underline{58.2} & \textbf{57.86} & \textbf{46.99} & \underline{42.62} & 49.49 & 50.13 & \underline{30.41} & \underline{33.88} & \textbf{37.65} & \textbf{35.08} & \textbf{47.05}
    \\
\hline
\end{tabular}
\end{adjustbox}
\label{retrieval_trumans}
\end{table*}
\begin{table*}[t]
\centering
\renewcommand{\arraystretch}{1.0} 
\small
\caption{Ablation Studies for \ours performed on HUMANISE+ \cite{wang2022humanise}. We report the average mRecall computed across ranks.}
\begin{adjustbox}{max width=\textwidth} 
\begin{tabular}{l|l|c|c|c|c|c|c||c|c|c|c|c|c||c}
\hline
\textbf{Protocol} & \textbf{Method} & \textit{st2m} & \textit{m2st} & \textit{ms2t} & \textit{t2sm} & \textit{tm2s} & \textit{s2mt} & \textit{t2m} & \textit{m2t} & \textit{s2m} & \textit{m2s} & \textit{t2s} & \textit{s2t} & \textit{avg.} \\
\hline
\multirow{4}{*}{All} & \ours             & \textbf{13.91} & \textbf{13.14} & \underline{8.46}  & \textbf{10.39} & \underline{4.09}  & \textbf{4.45} & 3.62  & 3.11  & \underline{1.12}  & \underline{1.16}  & 1.68  & \underline{2.51}  & \textbf{5.63}  \\
& - w/o cross-modal & 5.20           & 3.77           & \textbf{8.91} & 7.95          & 2.49  & 2.59          & \textbf{4.35} & \underline{3.21}  & 1.01  & 0.91  & \textbf{2.52} & \textbf{2.64} & 3.79  \\
& - w/o single      & \underline{11.91}          & \underline{12.93}          & 8.35  & \underline{8.96}          & \textbf{4.33} & \underline{4.31}          & 0.22  & 0.29  & 0.09  & 0.25  & 0.54  & 0.24  & \underline{4.36}  \\
& - w tri-modal          & 6.14           & 6.00           & 7.56  & 7.93          & 3.02  & 3.12         & \underline{4.16}  & \textbf{4.37} & \textbf{2.00} & \textbf{1.66} & \underline{2.10}  & 1.63  & 4.14 \\

\hline
\multirow{4}{*}{\makecell{Small \\ Batches}} & \ours             & \textbf{79.15} & \textbf{79.01} & 76.30 & \textbf{77.42} & \textbf{57.55} & \textbf{57.06} & \underline{63.20} & \underline{61.89} & \underline{34.11} & \underline{34.47} & \textbf{50.18} & \textbf{49.76} & \textbf{60.00} \\

& - w/o cross-modal & 65.47          & 63.26          & \textbf{78.11} & \underline{76.76}       & 49.89         & 49.31         & \textbf{63.98} & \textbf{62.76} & 32.23 & 32.52 & \underline{49.66} & \underline{48.95} & \underline{56.07} \\
& - w/o single      & \underline{75.31}          & \underline{76.21}          & \underline{76.42} & 75.50       & \underline{53.76}         & \underline{53.86}         & 15.71 & 16.15 & 12.03 & 14.56 & 17.02 & 14.82 & 41.77 \\
& - w tri-modal          & 62.48          & 62.46          & 69.02  & 68.29       & 48.33         & 48.20         &  61.42 & 60.00 & \textbf{36.57} & \textbf{37.15} & 45.80 & 44.70 & 53.70 \\
\hline
\end{tabular}
\end{adjustbox}
\label{tab:abl_humanise}
\end{table*}
\begin{table*}[t]
\centering
\renewcommand{\arraystretch}{1.0} 
\small 
\caption{Ablation Studies for \ours performed on TRUMANS+ \cite{jiang2024scaling}.  We report the average mRecall computed across ranks.}
\begin{adjustbox}{max width=\textwidth} 
\begin{tabular}{l|l|c|c|c|c|c|c||c|c|c|c|c|c||c}
\hline
\textbf{Protocol} & \textbf{Method} & \textit{st2m} & \textit{m2st} & \textit{ms2t} & \textit{t2sm} & \textit{tm2s} & \textit{s2mt} & \textit{t2m} & \textit{m2t} & \textit{s2m} & \textit{m2s} & \textit{t2s} & \textit{s2t} & \textit{avg.} \\
\hline

\multirow{4}{*}{All} & \ours & \underline{10.54} & \underline{10.29} & \underline{8.59} & \textbf{8.51} & \underline{12.71} & \underline{3.94} & 5.67 & 5.9 & 1.94 & 7.88 & \textbf{10.35} & \textbf{2.83} & \textbf{7.42} \\
& - w/o cross-modal & 5.89 & 6.26 & 6.58 & 6.3 & 12.01 & 3.1 & \textbf{6.95} & \textbf{6.84} & \underline{1.98} & \textbf{9.05} & \underline{8.26} & 2.06 & \underline{6.27} \\
& - w/o single & \textbf{12.95} & \textbf{13.01} & \textbf{7.01} & 6.28 & \textbf{12.99} & \textbf{4.76} & 0.19 & 0.06 & 0.49 & 3.67 & 2.71 & 0.07 & 5.34 \\
& - w tri-modal & 5.62 & 4.91 & 6.24 & \underline{6.34} & 10.29 & 3.05 & \underline{6.31} & \underline{6.3} & \textbf{2.04} & \underline{8.36} & 8.19 & \underline{2.58} & 6.02\\
\hline
\multirow{4}{*}{\makecell{Small \\ Batches}} 

& \ours & \underline{60.8} & \underline{61.4} & \textbf{58.2} & \textbf{57.86} & \underline{46.99} & \underline{42.62} & \underline{49.49} & \underline{50.13} & \underline{30.41} & \underline{33.88} & \textbf{37.65} & \textbf{35.08} & \textbf{47.04}\\

& - w/o cross-modal & 50.96 & 51.12 & 50.12 & 50.75 & 38.12 & 33.98 & \textbf{52.76} & \textbf{52.91} & 29.91 & 33.56 & 33.73 & 31.08 & 42.41 \\
& - w/o single & \textbf{66.96} & \textbf{68.43} & \underline{56.38} & \underline{56.46} & \textbf{51.59} & \textbf{48.38} & 12.5 & 12.43 & 16.84 & 20.86 & 15.9 & 12.18 & 36.57\\
& - w tri-modal & 47.35 & 48.72 & 50.77 & 49.61 & 41.4 & 36.48 & 48.89 & 49.24 & \textbf{31.33} & \textbf{35.23} & \underline{37.08} & \underline{33.52} & \underline{46.72} \\
\hline
\end{tabular}

\end{adjustbox}
\label{tab:abl_trumans}
\end{table*}

\subsection{Retrieval Tasks}

\noindent\textbf{Tasks and Metrics.} Given samples from three modalities, text ($t$), motion ($m$) and scene ($s$), we evaluate our model across several retrieval tasks: retrieving one modality given two others (\textit{st2m, ms2t, mt2s}), retrieving two modalities given one (\textit{m2st, t2ms, s2mt}), and retrieving one modality given another single modality (\textit{t2m, m2t, s2m, m2s, t2s, s2t}). 
For instance, the \textit{m2st} task requires finding the samples $s$ and $t$ which best suit the given $m$ across a pool of possible pairs of scenes and texts. 
For brevity, we refer to these tasks as double-to-single,  single-to-double, and single-to-single. 

Retrieval tasks are typically evaluated using Recall@$K$, where $K$ is a rank in \{1, 2, 3, 5, 10\}. 
This metric measures the percentage of times the correct sample is among the top $K$ results, ordered by the similarity scores provided by the model.   
However, due to space constraints, we omit detailed Recall results at all ranks, instead presenting the more concise Mean Recall (mRecall) from \cite{delmas2024poseembroider}, which represents the average Recall across ranks \{1, 2, 3, 5, 10\}. Comprehensive  Recall results are provided in the Supplementary Material.

\noindent\textbf{Protocols.} The evaluation adheres to the retrieval protocols outlined in \cite{petrovich23tmr, guo2022generating}. Specifically, the \commas All" protocol requires distinguishing correct modality samples across all samples within the test set, while the \commas Small Batches" protocol requires discrimination among batches of 32 samples.

\begin{figure}[ht!]
    \centering

    \begin{subfigure}{\linewidth}
        \begin{subfigure}{.32\linewidth}
            \centering
            \includegraphics[width=1\linewidth]{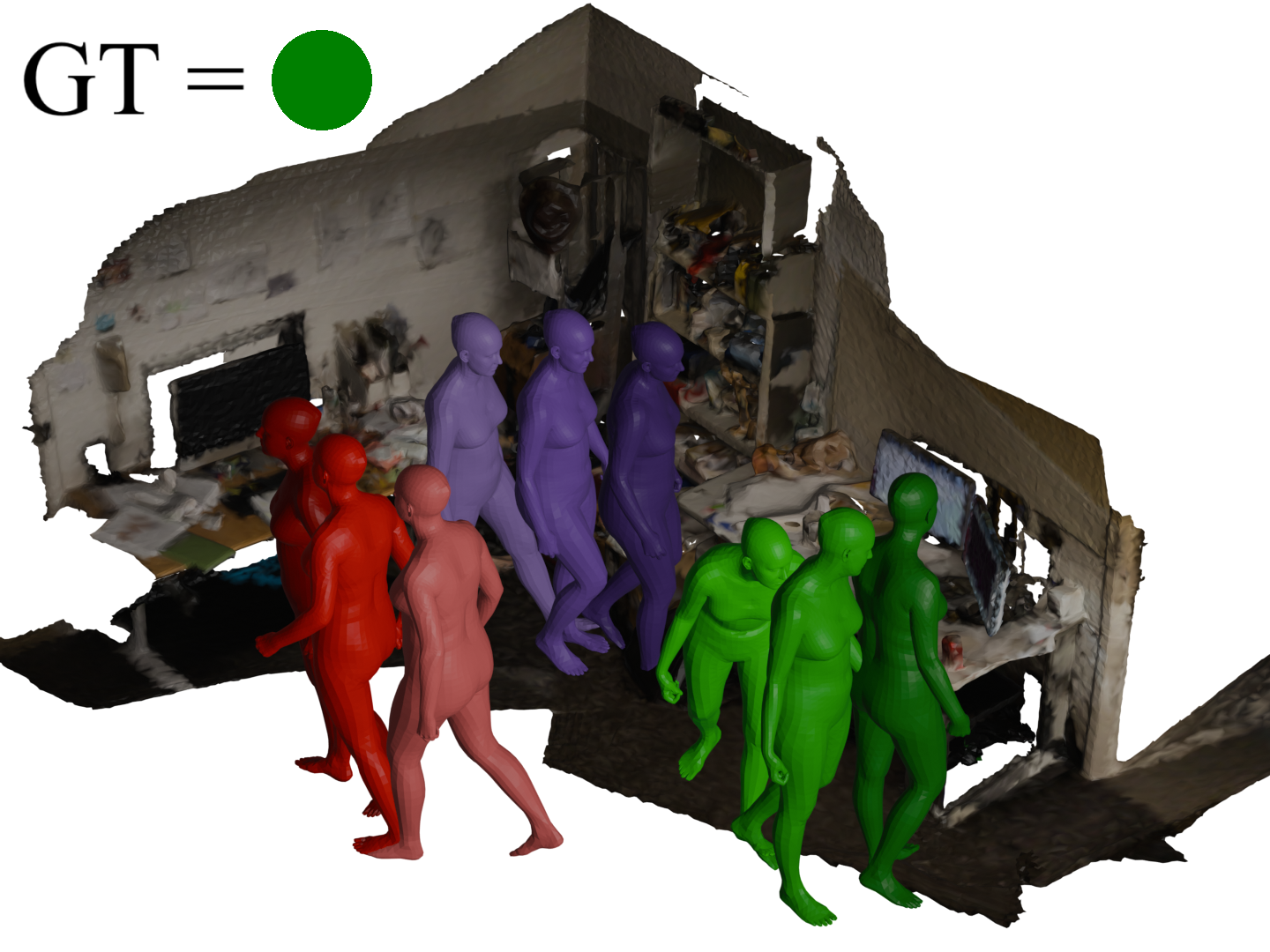}
            \caption{Walk to the desk near a computer and monitor, passing by a cabinet.}
            \label{fig:st2m-3}
        \end{subfigure}
        \hfill
        \begin{subfigure}{.32\linewidth}
            \centering
            \includegraphics[width=1\linewidth]{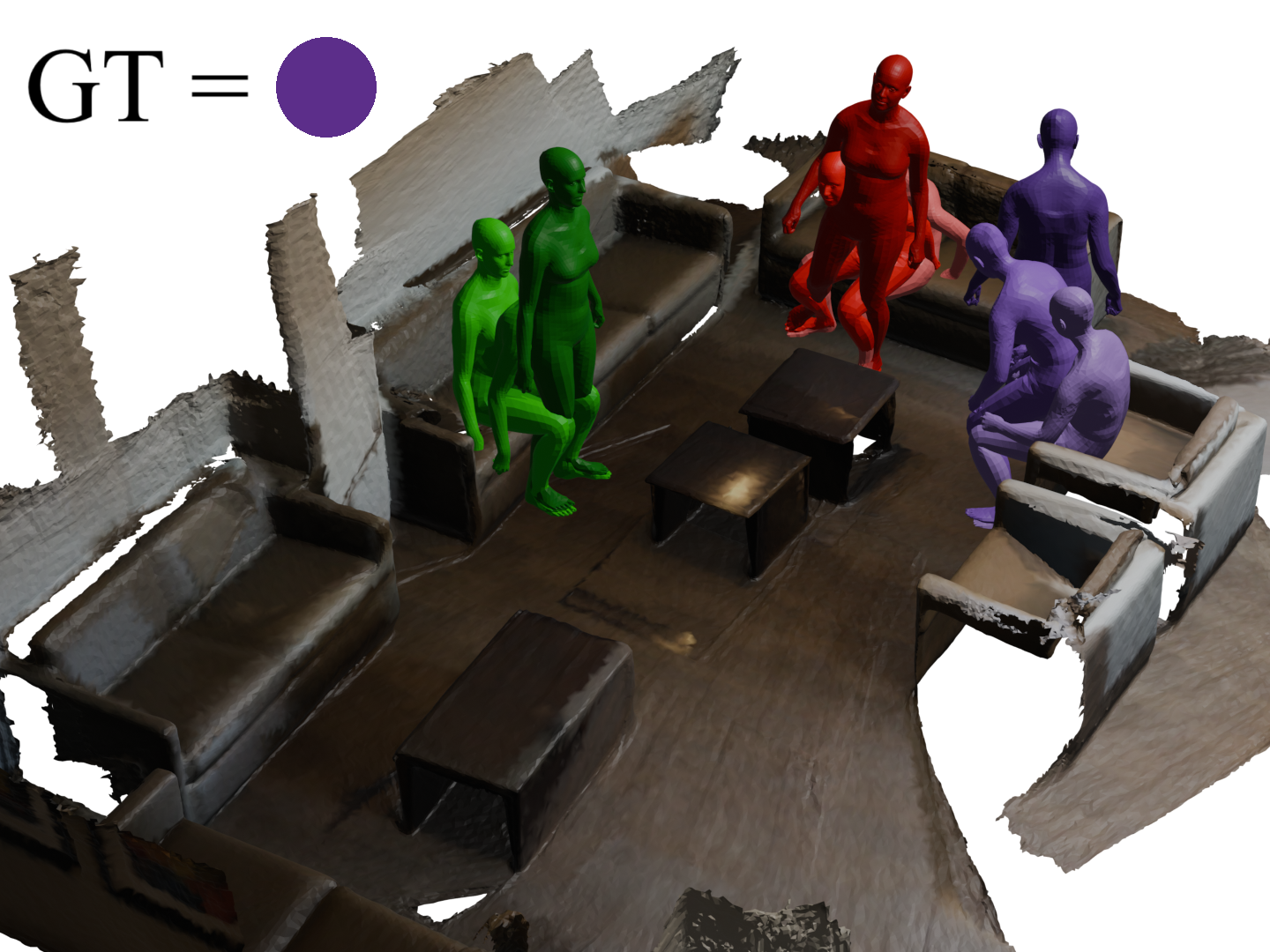}
            \caption{Stand up from the sofa chair, which is near the door, a table and a wall.}
            \label{fig:st2m-2}
        \end{subfigure}
        \hfill
        \begin{subfigure}{.32\linewidth}
            \centering
            \includegraphics[width=1\linewidth]{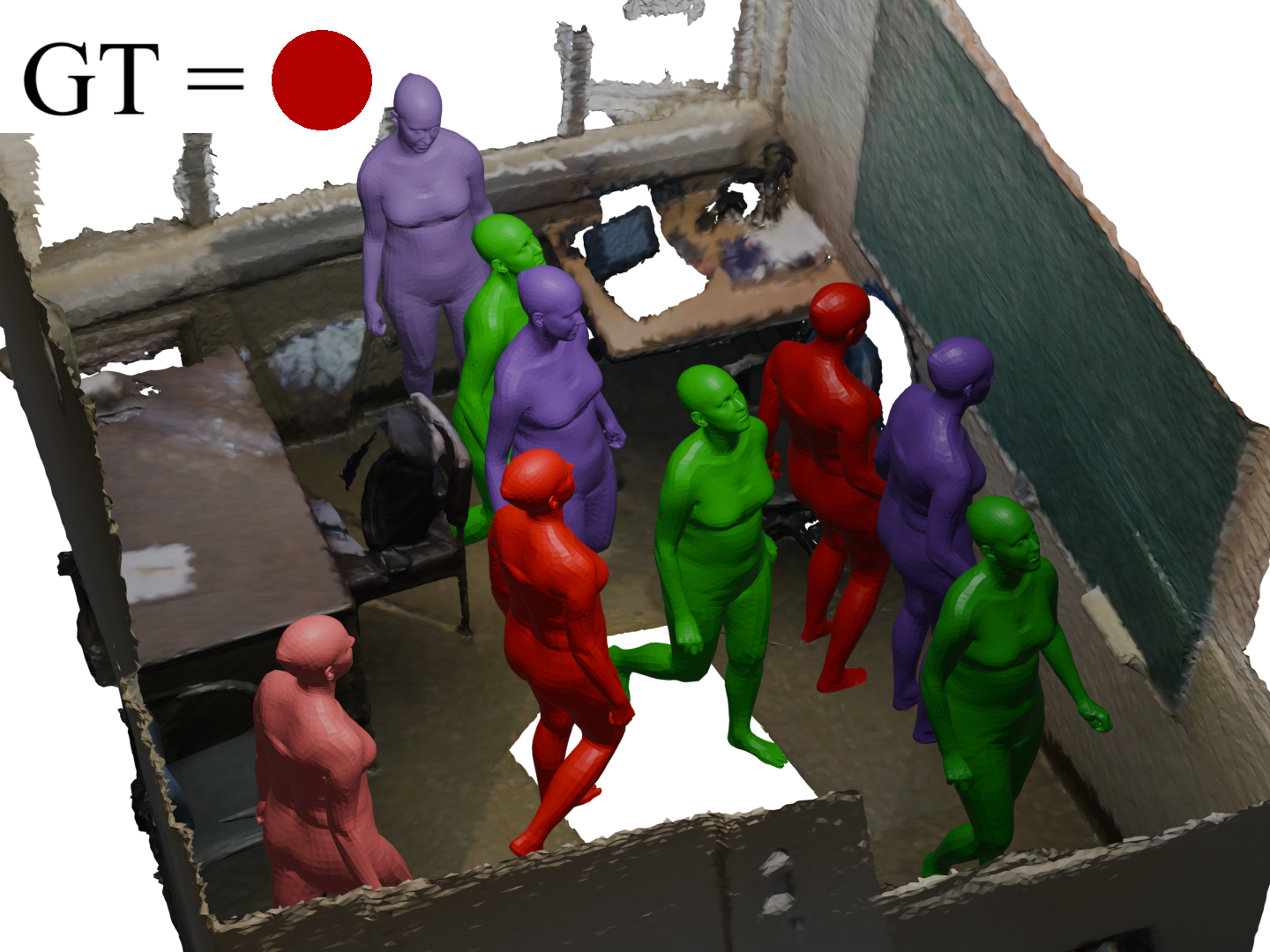}
            \caption{Walk to a chair in front of a laptop, passing by a backpack and a blackboard.}
            \label{fig:st2m-1}
        \end{subfigure}
    \end{subfigure}
    
    \begin{subfigure}{\linewidth}
        \begin{subfigure}{.32\linewidth}
            \centering
            \includegraphics[width=1\linewidth]{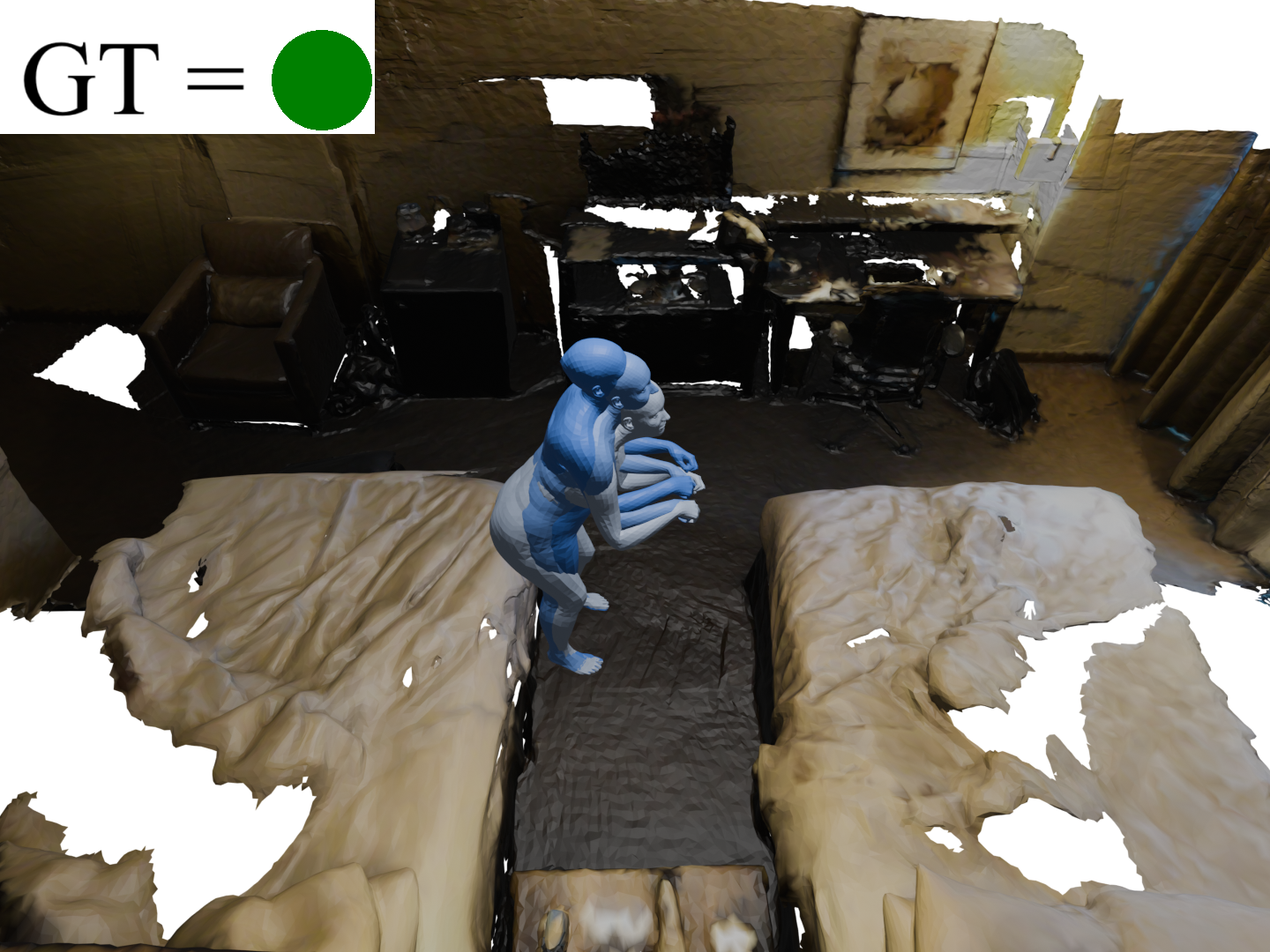}
            \caption{\textcolor{my_green}{\textbf{Stand up from the bed close to the armchair.}} \textcolor{my_purple}{Stand up from the bed, far from the office chair.} \textcolor{my_red}{Stand up from the bed near an armchair and a suitcase.}}
            \label{fig:sm2t-3}
        \end{subfigure}
        \hfill
        \begin{subfigure}{.32\linewidth}
            \centering
            \includegraphics[width=1\linewidth]{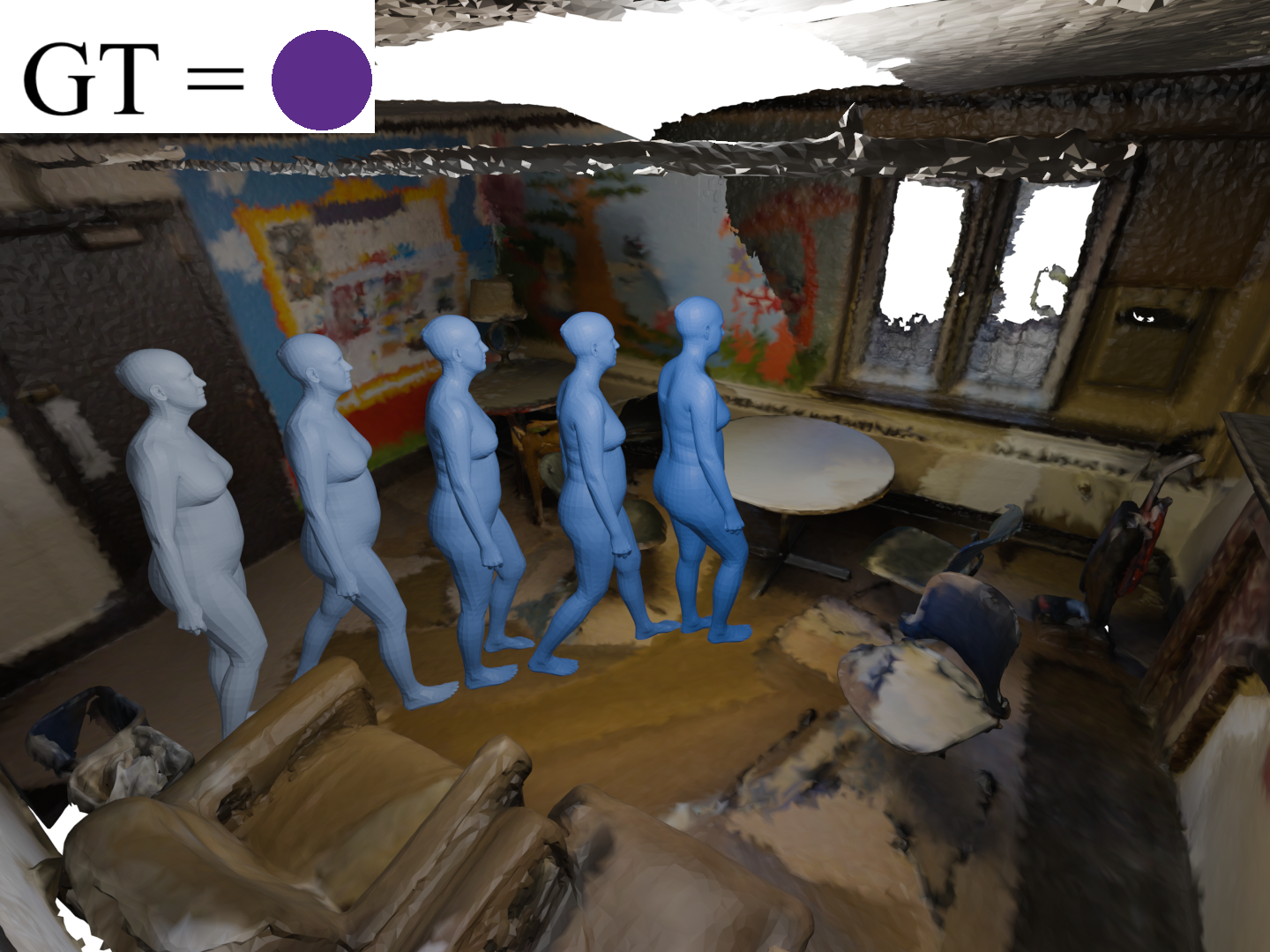}
            \caption{\textcolor{my_green}{Walk past an armchair to the table under the window.} \textcolor{my_purple}{\textbf{Walk to the table near the fireplace and a chair.}} \textcolor{my_red}{Walk past a table to the chair near the cabinet.}}
            \label{fig:sm2t-2}
        \end{subfigure}
        \hfill
        \begin{subfigure}{.32\linewidth}
            \centering
            \includegraphics[width=1\linewidth]{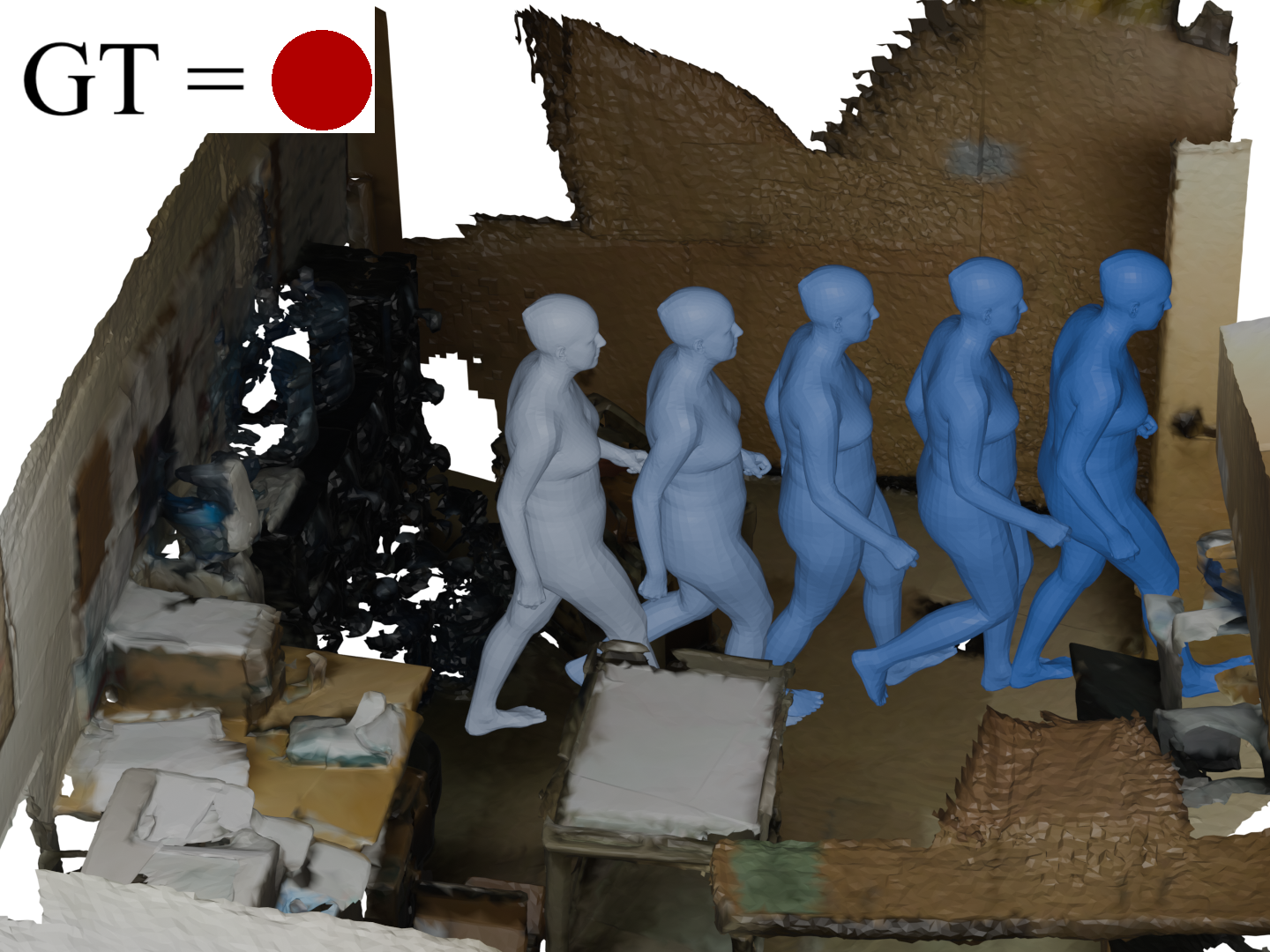}
            \caption{\textcolor{my_green}{Walk past the wall and extinguisher to the doorframe.}
            \textcolor{my_purple}{Walk to the doorframe, passing by an exit sign.}
            \textcolor{my_red}{\textbf{Walk to the door, passing by a table and a cabinet.}}}
            \label{fig:sm2t-1}
        \end{subfigure}
    \end{subfigure}
    \vspace{-0.5em}
    \caption{Qualitative examples for \textit{st2m} (\ref{fig:st2m-1}, \ref{fig:st2m-2}, \ref{fig:st2m-3}) and \textit{ms2t} (\ref{fig:sm2t-1}, \ref{fig:sm2t-2}, \ref{fig:sm2t-3}). \textcolor{my_green}{\textbf{First}}, \textcolor{my_purple}{\textbf{second}}, and \textcolor{my_red}{\textbf{third}} retrieved samples are shown. In each pictorial, GT=color \textit{(top-left corner)} indicates the correct corresponding motion \textit{(top row)} and text \textit{(bottom row)}.}
    \label{fig:reb-qualitatives}
\end{figure}

\noindent\textbf{Retrieval Results.}
In Table~\ref{tab:retrieval_humanise} and Table~\ref{retrieval_trumans}, we present \ours retrieval results. Since no other model has previously addressed retrieval across scene, text, and motion, we compare it with two state-of-the-art text-to-motion retrieval models, TMR \cite{petrovich23tmr} and MoPa \cite{yu2024exploring}. We report \textit{t2m} and \textit{m2t} as proxies for the \textit{st2m} and \textit{m2st} tasks and grey out these results, as they provide the extra input $s$ which neither TMR nor MoPa can use. To ensure a fair comparison (TMR+S and
MoPa+S), we start from their pretrained architecture and extend it with the same scene encoder as \ours, finetuning it on the target dataset. As they do not leverage cross-modal encoders we average their unimodal representations to produce cross-modal ones \cite{girdhar2023imagebind}. 

In the \textit{t2m} and \textit{m2t} tasks, \ours consistently outperforms all baselines on HUMANISE+, while TMR leads on TRUMANS+. We attribute this discrepancy to the limited value of scene cues in TRUMANS+: individual scenes often support several distinct motions. The dataset’s high motion diversity also makes text and motion signals more discriminative than scene context, even after recaptioning. TMR capitalizes on this with Guo features \cite{guo2022generating}, which ignore spatial ties between scene and motion and thus make tasks other than \textit{t2m}/\textit{m2t} impossible. TMR accuracy drops sharply without them (see Supplementary Material). By contrast, HUMANISE+ contains similar text–motion pairs, so scene information is essential for disambiguation. Note that this setting is the closest match to our target application of evaluating Human-Scene Interaction generations.

Overall, \ours emerges as the top performer for the majority of double-to-single and single-to-double tasks. 
It improves \commas All'' protocol's \textit{st2m} relative to the best among the evaluated models by 209\% on HUMANISE+.
On average, \ours outperforms the best scene-aware models by 76.47\% and 35.15\% on the \commas All'' protocol on HUMANISE+/TRUMANS+, and by 26.47\% and 10.44\% on \commas Small Batches'', showing superior multimodal alignment.

\noindent\textbf{Additional Results.}
In Fig.~\ref{fig:reb-qualitatives} we provide qualitative results of \ours's rankings for the \textit{st2m} and \textit{ms2t} tasks. This evaluation highlights both the complexity and challenging nature of these tasks, as many samples exhibit high semantic relevance to the conditioning modalities. Moreover, it demonstrates the effectiveness of \ours in retrieving motions that maintain a strong correspondence with the ground truth in its top-ranked results. More qualitative results are available in the Supplementary Material, including examples for other tasks, such as \textit{m2t} and \textit{mt2s}. 

Finally, we evaluate our model on HumanML3D \cite{guo2022generating}, a common text-to-motion dataset that does not include scene information. In Sec. 9 of the Supplementary Material we show that \ours achieves results comparable to state-of-the-art text-to-motion retrieval models \cite{petrovich23tmr, yu2024exploring}.

\subsection{Ablation Studies on the Model Design}
\label{sec:ablations}

In Table~\ref{tab:abl_humanise} and Table~\ref{tab:abl_trumans} we test ablative variants of \ours: the tested variants still have the same architecture but they differ by taking into consideration only unimodal or cross-modal relations in the loss. Recalling Sec.~\ref{sec:monster}, the ablation studies vary the set $K$ used for the contrastive terms.
In \commas \ours w/o cross-modal", we remove all cross-modal encoders, which also eliminates all cross-modal terms from the set $K$, resulting in the optimization of the set $\{(t,s), (m,t), (m,s)\}$. In this case we use text as the bridging modality to align all three modalities and average latents during inference to compute the cross-modal representations \cite{girdhar2023imagebind}. Predictably, this model has significant performance drops in most double-to-single and single-to-double retrieval tasks.
Next, in \commas \ours w/o single" we remove all terms involving only single-modality from $K$, resulting in optimizing the reduced set $\{(st,m), (mt,s), (ms,t)\}$. 
This model aligns single modalities' embedding to cross-modal embeddings only.
In some cases, the double-to-single retrieval task of the ablated model outperforms the corresponding performance of \ours on TRUMANS+. 
This improvement likely stems from a closer alignment between training and test objectives in these instances. 
However, on average, this model’s performance appears diminished compared to that of \ours (see \textit{avg.} in Table~\ref{tab:abl_humanise} and Table~\ref{tab:abl_trumans}). Indeed single-to-single retrieval performance shows a significant drop, highlighting the limitations of the ablated approach.
In addition to these ablation we also evaluate \commas{}\ours w tri-modal", a single tri-modal architecture inspired by \cite{delmas2024poseembroider}. Here, all modalities are embedded by the same tri-modal encoder and decoded jointly to compute cross-modal embeddings. While this model performs well on certain single-to-single retrieval tasks, it underperforms on double-to-single and single to double tasks. 
These experiments confirm that superior performance can be achieved by modeling higher-order relationships through the alignment of both unimodal-to-unimodal and unimodal-to-cross-modal interactions while retaining the rich feature semantics of the single encoders.

\subsection{MonSTeR for Evaluation}
In this section, we introduce \ours as a tool to assess the quality of text-conditioned Human Scene Interaction generative models (HSI). After testing its motion-scene grounding properties, we validate the alignment of \ours's scores with human judgment.

\noindent\textbf{MonSTeR Metrics.} FID and Recall@\{1,2,3\} are commonly employed in the evaluation of text-to-motion models  \cite{chen2023executingcommandsmotiondiffusion, guo2022generating, guo2023momaskgenerativemaskedmodeling, tevet2022humanmotiondiffusionmodel}. However, these metrics do not account for the scene and its interactions with text and motion, rendering them less suited for evaluating HSI models. By leveraging \ours's latent space, we can derive more informative versions of these metrics.

We extend FID to assess both the motions' plausibility and their coherence with the scene leveraging the embeddings from the cross-modal encoder $MS$ of \ours (cf. Fig.~\ref{fig:architecture}).  
We extend the Recall \cite{guo2022generating}, by using the \textit{m2st} task embeddings to measure the adherence of a motion to a text and a scene.

\noindent\textbf{Evaluating Path Plausibility.} To evaluate the scene-motion grounding capabilities of \ours, we assess if its latent space can  
differentiate between motions that adhere to paths coherent with the provided text and scene affordances, and those that do not.
To this end, we rotate motions from the test sets of \cite{wang2022humanise, jiang2024scaling}, starting at an angle of $0$ radians and increasing incrementally up to $\pi$, with the rotation pivot at the end position of the original motion. We ensure that each rotated motion stays within the scene boundaries and avoids interpenetrating any scene objects.

\begin{figure*}[h!]
    \centering
    \begin{subfigure}[b]{0.49\textwidth}
        \centering
        
        \includegraphics[width=0.85\textwidth]{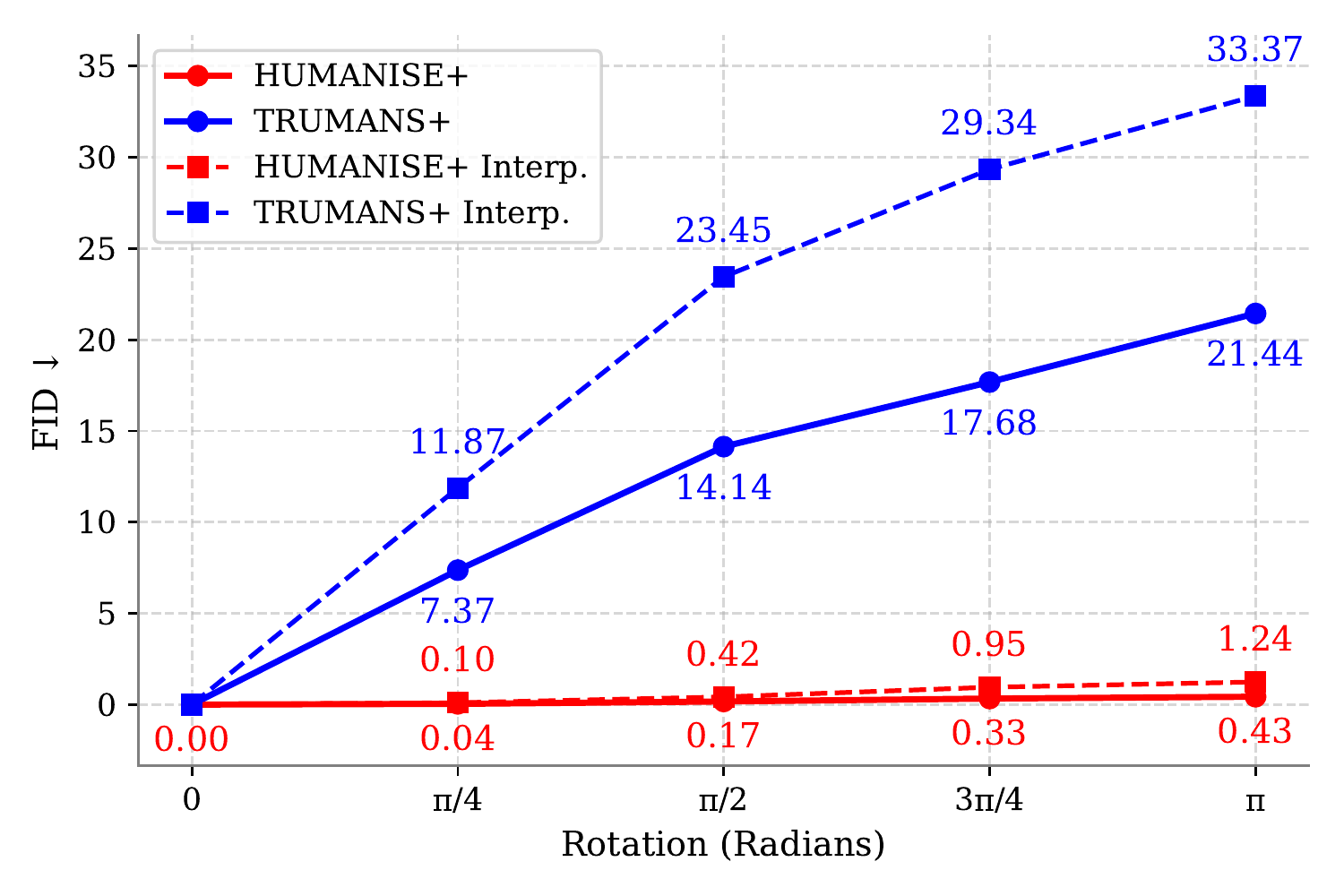}
        \caption{}
        \label{fig:image_a}
    \end{subfigure}
    \hfill
    \begin{subfigure}[b]{0.49\textwidth}
        \centering
        \includegraphics[width=0.85\textwidth]{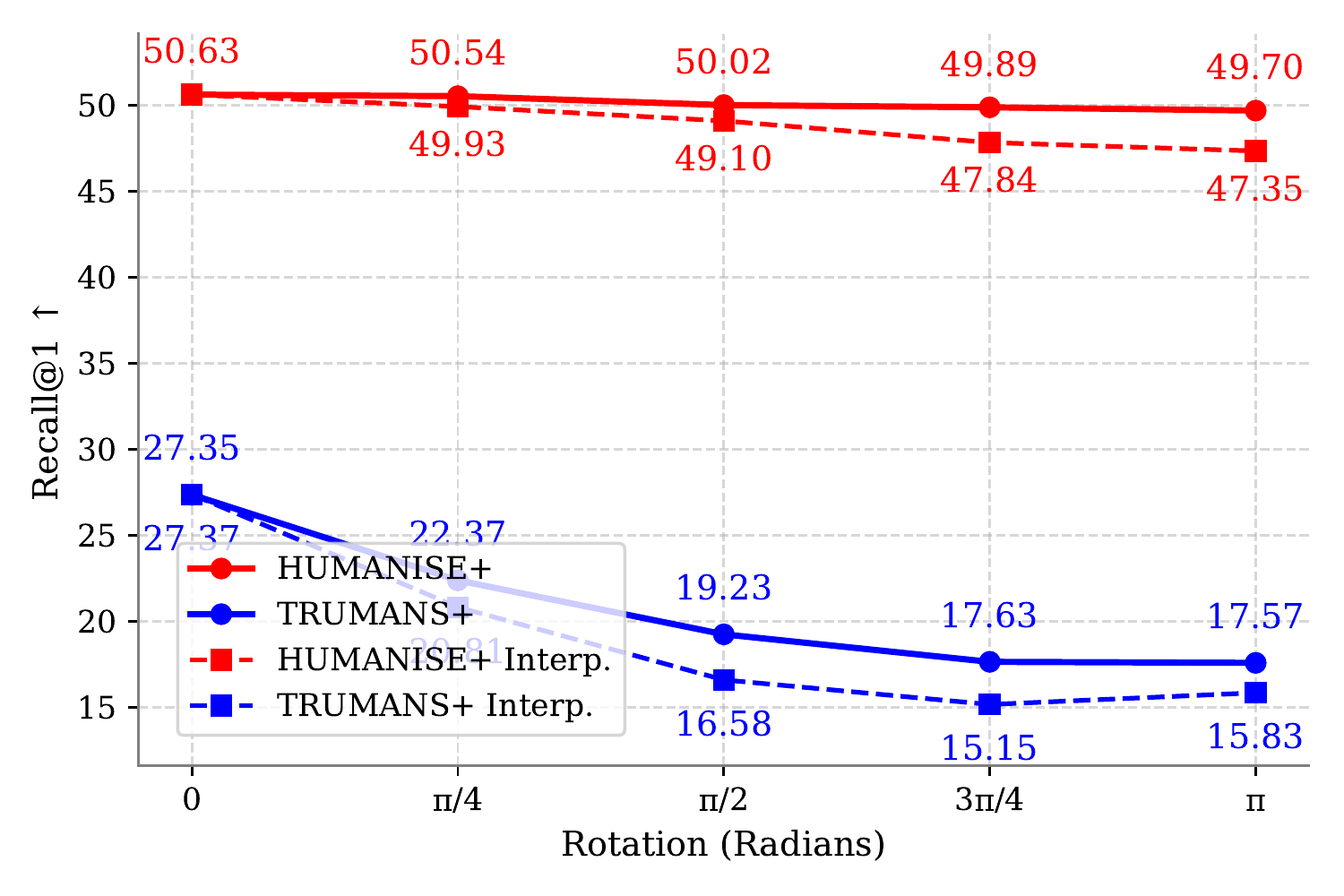} 
        \caption{}
        \label{fig:image_b}
    \end{subfigure}
    \caption{ \ours 's FID $\downarrow$ (a) and Recall@1 $\uparrow$ (b) trends when motions are increasingly rotated, from $0$ to $\pi$ radians.}
    \label{fig:plots}
\end{figure*}

Solid lines in Fig.~\ref{fig:plots} show that both FID and Recall values computed using \ours embeddings deteriorate as the rotation increases. 
This is both expected and desirable, as increased rotation results in a larger divergence between the original and modified paths, leading to a loss of coherence between the motion, the scene, and the conditioning text.
This is more evident in \cite{jiang2024scaling} due to the precise descriptions of contact interactions with small objects (e.g.\ \textit{pick up the keyboard on the desk});
when motions are rotated, the coherence is disrupted. 
The gap in Fig.~\ref{fig:plots} reflects datasets' properties.
\textsc{TRUMANS+} mixes dynamic motions with small targets in wide spaces, so small rotations quickly misalign modalities and lower scores.
\textsc{HUMANISE+}, dominated by static actions around large, nearby objects, is far less sensitive to the same rotations.

\noindent\textbf{Evaluating Compliance with the Scene.} We repeat the experiment above but enable motion rotations that would lead to interpenetration. In Fig.~\ref{fig:plots}, comparing the dotted lines to the solid ones, we see that including motions penetrating scene objects causes an even further decline in these metrics. This suggests that the latent space of \ours has also internalized that natural motions should not involve interpenetration with the scene.

\noindent\textbf{Alignment to Human Preferences.}
We perform a user study to test the alignment of \ours's ranking with human preferences. We generate motions using two state-of-the-art models for text-conditioned human-scene interaction: the conditional variational auto-encoder from \cite{wang2022humanise} and the two-stage diffusion model from \cite{wang2024move}. 
For each pair of generated motions, we rank them using \textit{st2m} score and compare this ranking to human preferences. We collect 1122 annotations from 224 evaluators. The results show that \ours's rankings align with human evaluations 66.5\% of the time, indicating agreement with human judgment.

\subsection{Downstream Tasks}

\noindent\textbf{In Scene Object Placement.}
We question if \ours latent space can discriminate if an object is placed on the correct $x,y,z$ coordinates.
To this aim, we extrapolate objects of interest, and set up a $5\times5\times5$ grid around their ground-truth positions. We shift the object into each of the 125 cells in the grid, creating an equal number of modified scenes.

Then, we leverage the \textit{mt2s} score of \ours to evaluate the similarity between the text-motion embedding and each scene variant.
This process provides a score for each pairing $ ((\text{text},\text{motion}),\text{scene}_i), \text{with } i \in [1,125].$

We locate the object of interest by selecting the highest score provided by our model and compute the $L2$ distance to its real position.
As we use $25\times25\times25$ cm cells the error ranges in $[0\text{ cm}, 86.6\text{ cm}]$, with the maximum being twice the length of a cell's diagonal. While the average error is 58.98 cm, our zero-shot prediction scores only 18cm.
\ours can accurately localize objects in a zero-shot setting, leveraging spatial knowledge grounded in the text-motion latent representations. Further details and visualizations are available in Sec.~5 of the Supplementary Material.

\vspace{-1em}
\paragraph{Motion Captioning.} To test the descriptiveness of \ours's latent representations, we use its embeddings for the downstream task of Motion Captioning. In particular, we freeze \ours's latent space and feed its motion embeddings to GPT2 \cite{radford2019language} which we then train to caption motions. We then compare our results with MotionGPT \cite{jiang2024motiongpt}, retrained from scratch on both HUMANISE+ and \mbox{TRUMANS+}, adopting the evaluation protocol from \cite{jiang2024motiongpt}.

As shown by Table~\ref{tab:hum-capt} and Table~\ref{tab:trum-capt}, \ours is able to surpass MotionGPT on most metrics by a significant margin. Intuitively, the motion embedding from our model carries more semantic information as it has previously been aligned with the scene. Moreover, the higher ROUGE$_L$ and Bert$_{F1}$ suggest that the model not only predicts captions that preserve the word ordering of the ground truth but also generates descriptions that are closer to natural human language.

\begin{table}[ht!]
    \centering
        \begin{adjustbox}{width=\linewidth} 
        \begin{tabular}{c|c|c|c|c|c}
            \hline
             Method & BLEU 1 & BLEU 4 & ROUGE L & CIDER & BERT F1 \\
             \hline
             mGPT  & 42.16 & 17.47  & 40.23  & 11.13 & 22.16  \\
             \ours + GPT2  & \textbf{42.93} & \textbf{23.59} & \textbf{50.85} & \textbf{13.70} & \textbf{35.57} \\
             \hline
        \end{tabular}
        \end{adjustbox}
        \caption{Captioning performances on HUMANISE+ \cite{wang2022humanise}.}
    \label{tab:hum-capt}
\end{table}
\vspace{-2em}

\begin{table}[ht!]
    \centering
        \begin{adjustbox}{width=\linewidth} 
        \begin{tabular}{c|c|c|c|c|c}
            \hline
             Method & BLEU 1 & BLEU 4 & ROUGE L & CIDER & BERT F1 \\
             \hline
             mGPT  & 39.62  & 18.13  & 40.62 & \textbf{15.08} & 17.59   \\
             \ours + GPT2 & \textbf{42.82} & \textbf{21.59} & \textbf{45.98} & 12.24 & \textbf{26.66} \\
             \hline
        \end{tabular}
        \end{adjustbox}
        \caption{Captioning performances on TRUMANS+ \cite{jiang2024scaling}.}
    \label{tab:trum-capt}
\end{table}
\section{Limitations}

We train cross-modal encoders only on aligned modality pairs, due to the unfeasible computational expense of training on unpaired data.
Also, our scenes are static, meaning human actions do not alter the scene layout. While only a few works explore dynamic scene encoders, this remains a promising direction for future extensions of \ours.

\section{Conclusion}
\label{sec:Conclusion}

We introduced a novel retrieval model that unifies text, motion, and scene within a shared latent space. Inspired by topological deep learning, \ours captures higher-order interactions across all three modalities. We validated it through retrieval tasks, demonstrating the effectiveness of our unified representation. Through comprehensive ablation studies, we highlighted the contribution of each component to the model's overall performance.
Also, we proposed \ours as a tool for grounding Human-Scene Interaction models within the scene context and evaluated its alignment with human preferences. We applied our model on in-Scene Object Placement and Motion Captioning, showing its ability to generalize beyond traditional retrieval scenarios.

\section{Acknowledgement}
We acknowledge partial financial support from Wsense S.r.l., from the PNRR MUR project PE0000013-FAIR (CUP: B53C22003980006), from the Sapienza grants RG123188B3EF6A80 (CENTS), and RM1241910E01F571 (V3LI). This work has been carried out while M.P. was enrolled in the Italian National Doctorate on Artificial Intelligence run by Sapienza University of Rome, funded by the European Union – Next Generation EU, Mission 4 Component 1 CUP B53C22003870006. O.L is a Taub fellow and is supported by the Azrieli Foundation Early Career Faculty Fellowship. He is also supported by the Israel Science Foundation through a personal grant (ISF 624/25) and an equipment grant (ISF 2903/25).
{
    \small
    \bibliographystyle{ieeenat_fullname}
    \bibliography{main}
}

\end{document}